\documentclass[journal]{IEEEtran}
\usepackage[table]{xcolor}
\usepackage{multirow}
\usepackage{booktabs}
\usepackage{amssymb}
\usepackage[cmex10]{amsmath}
\usepackage{isomath}
\usepackage{graphicx}
\usepackage{xcolor}
\usepackage{algorithm}
\usepackage{algorithmic}
\usepackage{cite}
\graphicspath{{./Figures/}}
\usepackage{booktabs} 
\usepackage{enumitem}  
\newcommand{\bL}{\boldsymbol{\Lambda}}				
\newcommand{\re}{\text{e}}										
\newcommand{\ri}{\text{i}}										
\newcommand{\rFL}{\text{\tiny{FL}}}	  				
\newcommand{\rL}{\text{\tiny{L}}}							
\newcommand{\tH}{^{\text{H}}} 								
\newcommand{\tT}{^{\text{T}}} 								
\DeclareMathOperator{\E}{E}								
\DeclareMathOperator{\diag}{diag}								
\begin{document}
%
\title{A New Class of Efficient Adaptive Filters for Online Nonlinear Modeling}
%
\author{Danilo~Comminiello,~\IEEEmembership{Senior~Member,~IEEE,}
        Alireza~Nezamdoust, Simone~Scardapane, 
        Michele~Scarpiniti,~\IEEEmembership{Senior~Member,~IEEE,}
        Amir~Hussain,
        and~Aurelio~Uncini,~\IEEEmembership{Member,~IEEE}
\thanks{D. Comminiello, A. Nezamdoust, S. Scardapane, M. Scarpiniti and A. Uncini are with the Department of Information Engineering, Electronics and Telecommunications (DIET), Sapienza University of Rome, Via Eudossiana 18, 00184 Rome, Italy.}
\thanks{A. Hussain is with the School of Computing, Edinburgh Napier University, 10 Colinton Road EH10 5DT, Scotland, UK.}
}
%
%
%
\maketitle
%
%
%
%
%
\begin{abstract}
Nonlinear models are known to provide excellent performance in real-world applications that often operate in non-ideal conditions. However, such applications often require online processing to be performed with limited computational resources. To address this problem, we propose a new class of efficient nonlinear models for online applications. The proposed algorithms are based on linear-in-the-parameters (LIP) nonlinear filters using functional link expansions. In order to make this class of functional link adaptive filters (FLAFs) efficient, we propose low-complexity expansions and frequency-domain adaptation of the parameters. Among this family of algorithms, we also define the partitioned-block frequency-domain FLAF, whose implementation is particularly suitable for online nonlinear modeling problems. We assess and compare frequency-domain FLAFs with different expansions providing the best possible tradeoff between performance and computational complexity. Experimental results prove that the proposed algorithms can be considered as an efficient and effective solution for online applications, such as the acoustic echo cancellation, even in the presence of adverse nonlinear conditions and with limited availability of computational resources.
\end{abstract}
%
\IEEEpeerreviewmaketitle
%
%
%
%
%
\section{Introduction}
\label{sec:intro}
\IEEEPARstart{E}{mploying} nonlinear adaptive learning models in unknown environments has always been a relevant research topic in various signal processing applications aiming at tracking the statistical behavior of input signals \cite{ComminielloALM2018, ZhangTSMC2018, PanTSMC2018, ChengCOGN2021, LiuTSMC2021, MaCOGN2022}. Their ease of low-cost implementation still makes the development of new nonlinear filters very interesting for online applications working in real time or having significant computational constraints. In particular, among the various classes of filters, the so-called linear-in-the-parameters (LIP) nonlinear adaptive filters have been widely utilized due to their capabilities in many fields of application \cite{ComminielloALM2018, Sicuranza2009, CariniSIGPRO2013}, including audio and speech processing applications, e.g., \cite{PatelEUSIPCO2016, ComminielloWIRN2017}. 

One of the most popular families of LIP adaptive nonlinear filters is the one based on functional links, which is known in the literature as functional link artificial neural network \cite{PatraTSMC1999, SicuranzaTASL2011, zhao2008functional}, functional link network \cite{pao1994learning, PatelTCASI2016}, or functional link adaptive filter (FLAF) \cite{ComminielloTASL2013}. The FLAF model has been widely employed in recent years due to its efficient and flexible architecture based on a Hammerstein scheme \cite{PatraTSMC1999, SicuranzaTASL2011, comminiello2011functional, ComminielloTASL2013, comminiello2014nonlinear, ComminielloNEUNET2015, PatelTCASI2016, CariniICASSP2017, comminiello2017combined, ZhangCIM2017, ZhangTNNLS2018, ComminielloISCAS2018, ColaceNEUCOM2020, NayakaBSPC2020}. The analysis of FLAF modeling performance has also been conducted for speech and audio signal processing in various works \cite{ComminielloTASL2013, ComminielloIJCNN2015}. In \cite{ComminielloTASL2013}, the FLAF is introduced as a useful method for enhancing the original signal in the input side. This enhancement is achieved by representing the input signals in a higher dimensional space. A flexible split FLAF nonlinear adaptive filtering approach is introduced in \cite{comminiello2014nonlinear} for both linear and nonlinear coefficient adaptation. In \cite{comminiello2014nonlinear}, a proportionate FLAF method is studied and it has been shown that it may be preferred over the split FLAF in several nonlinear applications. The combination of adaptive filters on the nonlinear path of the FLAF architecture is carried out in \cite{ComminielloNEUNET2015, PatelEUSIPCO2017, comminiello2017combined}, where it has been proved to provide a higher performance against strong nonlinearities.

Most of the LIP nonlinear filters are adapted by using time-domain adaptive filters
. However, employing time-domain adaptive filters in real-time signal processing applications has the drawback of requiring extensive computational resources, which are often proportional to the length of the filters, which may involve even thousands of coefficients depending on the application. Over the years, many solutions have been adopted to mitigate this problem. Mini-batch adaptive filters were introduced in \cite{clark1983unified} employing a periodic update rule in order to reduce the computational cost. However, the most significant progresses in that sense have been obtained by using frequency-domain adaptive filters (FDAFs), which have been proved to be a powerful tool in processing signals for different applications.

With respect to time-domain algorithms, FDAFs show good convergence performance even in real-time applications. However, the input-output delay has always represented a problem \cite{clark1983unified, dentino1978adaptive, soo1990multidelay}. Early frequency-domain adaptive algorithms were more focused on reducing the computational complexity. In such problems, preserving the resources was preferred over the performance improvement \cite{narayan1981frequency, mansour1982unconstrained, mikhael1988fast}. Regardless of notable advancements in computational resources, the study of the frequency-domain methods is still being given considerable attention. The frequency-domain filtering approach has been employed in many online applications, such as in \cite{shynk1992frequency, buchner2005generalized, KuechTSP2005, HerbordtTASL2007, DietzenIWAENC2016, BernardiTASL2017, ComminielloICASSP2019}. 
Among the frequency-domain algorithms, the partitioned-block FDAFs have been used frequently in recent years for low-complexity applications due to their favorable properties of reducing both the input delay and the computational cost \cite{ChanTSP2001, LeeSPL2013, ValeroSPL2016, BernardiTASL2017, YangEUSIPCO2019}.
Despite the numerous developments of linear models in the frequency domain, the applicability of FLAFs in the frequency domain is still to be addressed, mainly due to some challenges, such as the frequency-domain adaptation after the functional transformation. 

Motivated by the above considerations, we design a class of frequency-domain FLAFs for online nonlinear modeling problems with the aim of providing low computational complexity and high-performance results. These goals are achieved through two important features of this class of filters. 
\begin{enumerate}[label=\roman*)]
	\item The first insight concerns the adaptation algorithm of the nonlinear filter, which is developed considering an implementation in the frequency domain. In particular, we derive the novel partitioned-block frequency-domain FLAF, which shows the best compromise between computational costs and performance, while reducing as much as possible any input-output delay. 
	\item The second relevant insight relies on developing an efficient strategy of nonlinear expansion. In particular, we provide the best low-complexity functional expansion suitable for the frequency-domain implementation. The proposed family of algorithms exploits the efficiency of frequency-domain adaptive filtering while benefiting from the effectiveness of the nonlinear estimation using functional links.
\end{enumerate}

\noindent The main advantage of the proposed family of algorithms is that it is able to provide effective performance results in online applications with low computational demand and also under adverse nonlinear conditions. This result is not so obvious when nonlinear filters are used because they require an accurate setting that depends very much on the scenario in which they are used, i.e., on how strong is the nonlinearity to be modeled and how many resources are available for processing. Our approach is able to provide superior modeling performance even when conditions change dramatically. To this end, we have assess extensively the capabilities of the proposed frequency-domain FLAFs in nonlinear acoustic echo cancellation (NAEC) scenarios. 

Overall, we summarize the paper contributions as follows:
\begin{itemize}
    \item We propose a novel general framework of linear-in-the-parameters nonlinear adaptive filters for online applications involving low-cost nonlinear expansion and efficient learning.
    \item We provide a new perspective on different functional link expansions from the point of view of the computational complexity and we provide the best possible functional link expansion considering a tradeoff between performance and complexity.
    \item We define a novel partitioned-block frequency-domain FLAF as an efficient solution for online applications and we discuss its performance analysis.
    \item We evaluate the proposed algorithms in a classic online application, that is the nonlinear acoustic echo cancellation, considering both stationary and nonstationary conditions, as well as different signal distortion and magnitude levels.
\end{itemize}

The rest of the paper is organized as follows. Section \ref{sec:flaf} introduces the problem formulation and the methodological background, including the nonlinear FLAF model and its application to online problems like nonlinear acoustic echo cancellation. In Section \ref{sec:feb}, we describe the main functional expansions, providing a perspective on their computational demands and how to design a low-cost implementation. The proposed family of low-cost FLAF in the frequency-domain is defined in Section \ref{sec:fdflaf}. We focus in particular on the proposal of a partitioned-block frequency-domain FLAF, whose performance analysis is discussed in Section \ref{sec:pbcomplexity}. In Section \ref{sec:results}, we describe the experimental settings and discuss the achieved results. Finally, Section \ref{sec:conc} concludes the paper.
%
%
%
%
%
\section{The Functional Link LIP Nonlinear Adaptive Filter Model and Its Applications to NAEC}
\label{sec:flaf}
%

\begin{figure}[t]
	\centering
	\includegraphics[width=0.98\columnwidth,keepaspectratio]{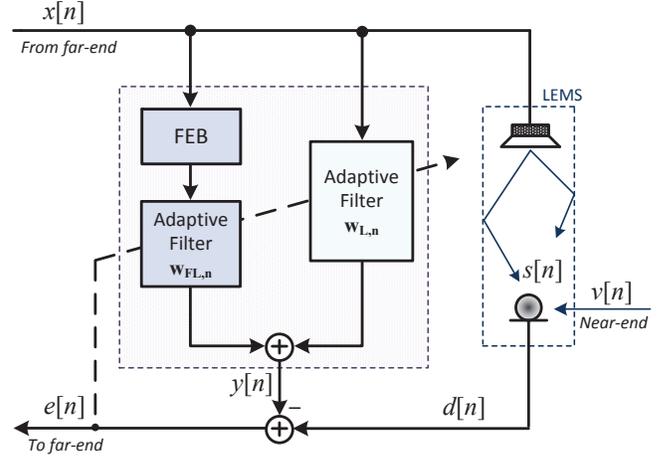}
	\caption{A split architecture of an LIP nonlinear filter implemented in a nonlinear acoustic echo cancellation scheme.}
	\label{fig:slip}
\end{figure}

In order to understand the reasons that underpin the family of proposed algorithms, in this section we provide a problem formulation illustrating the principles of the FLAFs and their application to practical online signal processing problems, such as in NAEC. In such scenarios, the system output $y\left[n\right]$ is a combination of linear and nonlinear components. For this reason, a parallel split filtering architecture, like the one shown in Fig.~\ref{fig:slip}, is one of the most appropriate choices as a canceler system. Here we consider a split functional link adaptive filter \cite{ComminielloTASL2013}, in which a nonlinear filter is implemented in parallel with a linear one. The latter filter aims at modeling the unknown linear part related to the acoustic impulse response. The nonlinear filter, instead, focuses only on the modeling of any nonlinear distortion, regardless of the estimation of linear components of the acoustic impulse response. The nonlinear filter is an LIP nonlinear adaptive filter involving a cascade of a functional expansion block and a linear filter.

Let us consider the online application of NAEC, represented in Fig.~\ref{fig:slip}. The signal $x\left[n\right]$ coming from the remote far-end communication user is reproduced by the loudspeaker and captured by the microphone on the near-end side. The path from the loudspeaker to the microphone is described by an acoustic impulse response, which is estimated by the linear adaptive filter. On the other hand, the loudspeaker introduces some nonlinear distortion that affects the reproduced signal. Such distortions are modeled by means of the nonlinear filter of the split FLAF. The distorted and reverberated signal, denoted by $s\left[n\right]$ and representing the echo signal to be estimated and canceled, is then captured by the microphone together with the environmental background noise $v\left[n\right]$. The resulting microphone signal $d\left[n\right] = s\left[n\right] + v\left[n\right]$ is also denoted as the desired signal. The near-end system, also known as loudspeaker enclosure-microphone system (LEMS), is the system to be modeled and contains a mixture of linear components, due to the acoustic path, and nonlinear components generated by the loudspeaker. 

The input signal reproduced by the loudspeaker is received as input by the split FLAF model, which passes it in the linear path as ${\mathbf{x}}_{\rL,n}  \in \mathbb{R} ^{M}  = \left[ {x\left[ n \right]} \right.$ $\left. {\begin{array}{*{20}c} {x\left[ {n - 1} \right]} &  \ldots  & {x\left[ {n - M + 1} \right]}  \\ \end{array}} \right]\tT \in\mathbb{R}^{M}$, where $M$ is the length of the input vector. The signal ${\mathbf{x}}_{\rL,n}$ is processed by the linear filter $\mathbf{w}_{\rL,n} \in \mathbb{R}^{M} = \left[ {{w_{\rL,0}\left[ n \right]}}\right.$ $\left.{{\begin{array}{*{20}l} {w_{\rL,1}\left[n \right]} &  \ldots  & {w_{\rFL,M_{\re}-1}\left[ {n} \right]} \\ \end{array}}} \right]\tT \in\mathbb{R}^{M}$ and provides the linear output. 
Concurrently, the input signal $x\left[n\right]$ is also processed by the nonlinear filter of the canceler whose input vector can be expressed as ${\mathbf{x}}_{\rFL,n}  \in \mathbb{R} ^{M_{\ri}}  = \left[ {\begin{array}{*{20}c} {x\left[ n \right]} & {x\left[ {n - 1} \right]} &  \ldots  & {x\left[ {n - M_{\ri} + 1} \right]}  \\ \end{array}} \right]\tT \in\mathbb{R}^{M_{\ri}}$, where its length $M_{\ri}$ may be smaller than $M$. Indeed, the acoustic impulse response may involve thousands of parameters to be estimated, while only a small part of its length may be sufficient for modeling the nonlinear distortion. 

The signal ${\mathbf{x}}_{\rFL,n}$ is nonlinearly expanded by the functional expansion block, which includes the set of functional links, denoted by ${\boldsymbol{\Phi}}=\{ \varphi_0(\cdot),\varphi_1(\cdot), \ldots,\varphi_{Q-1}(\cdot)\}$, where $ Q $ is the number of different functions of the set. The $ i $-th element of $ \mathbf{x}_{\rFL,n} $ is nonlinearly transformed by the functional link set, generating the vector $\overline{\mathbf{g}}_{i,n} = \left[ {\begin{array}{*{20}c} \varphi_0(x[n-i]) & \varphi_1(x[n-i]) & \ldots & \varphi_{Q-1}(x[n-i])\\ \end{array}} \right]\tT$ $\in$ $\mathbb{R}^{Q}$. Concatenating all the vectors $ \overline{\mathbf{g}}_{i,n} $, for $ i=0,\ldots,M_{\ri}-1 $ leads to the final nonlinear vector $ \mathbf{g}_n \in\mathbb{R}^{M_{\re}}$ as:
\begin{equation}
	\begin{split}
		\mathbf{g}_n &= \left[ {\begin{array}{*{20}c} {\mathbf{\overline{g}}_{0,n}\tT} & {\mathbf{\overline{g}}_{1,n}\tT} & \ldots & {\mathbf{\overline{g}}_{M_{\ri}-1,n}\tT} \end{array}} \right]\tT \\
		&= \left[\begin{array}{*{20}c} {g_{0}\left[n\right]} & {g_{1}\left[n\right]} & \ldots & {g_{M_{\re}-1}\left[n\right]} \end{array}\right]\tT,
	\end{split}
	\label{equ2}
\end{equation}

\noindent where the overall length is usually larger than the input vector length, i.e., $ M_{\re} \geq M_{\ri} $. The nonlinear vector $\mathbf{g}_n$ does not contain any linear elements and it is processed by $\mathbf{w}_{\rFL,n} \in \mathbb{R}^{M_{\re}} = \left[ {{\begin{array}{*{20}c} {w_{\rFL,0}\left[ n \right]} & {w_{\rFL,1}\left[n \right]} &  \ldots  & {w_{\rFL,M_{\re}-1}\left[ {n} \right]} \\ \end{array}}} \right]\tT$, thus producing the system output that estimates the distorted far-end signal. 

The input signals to the two parallel branches of the split FLAF can be written in a compact fashion, as well as the filter vectors \cite{ComminielloEUSIPCO2017}:
\begin{align}  \label{equ5}
	\mathbf{x}_n\in \mathbb{R}^{M+M_{\re}} = \left[
	\begin{matrix}
		\mathbf{x}_{\rL,n} \\
		\mathbf g_n\\
	\end{matrix}
\right],
\end{align}

\begin{align}  \label{equ6}
	\mathbf{w}_n\in \mathbb{R}^{M+M_{\re}} = \left[
	\begin{matrix}
		\mathbf{w}_{\rL,n} \\
		\mathbf{w}_{\rFL,n} \\
	\end{matrix}
	\right].
\end{align}

\noindent Therefore, the overall output of the model estimating the near-end LEMS signal can be derived in a joint approach as: 
\begin{align}  \label{equ4}
y\left[n\right] = \mathbf{x}_n\tT \mathbf{w}_{n-1}.
\end{align}

\noindent This signal can be subtracted from the microphone signal $d\left[n\right]$, thus obtaining the echo-free error signal $e \left[n\right] = d\left[n\right]-y\left[n\right]$ to be delivered to far-end communication user.

The filter updates can be performed by adopting any adaptive algorithm (see for example \cite{sayed2011adaptive}, \cite{uncini2015fundamentals}), also including sparse models or even more sophisticated filtering architectures \cite{ComminielloALM2018, ComminielloICASSP2019, ScardapaneNEUCOM2017, ComminielloNEUNET2015, comminiello2017combined}. In this work, in order to achieve a low-complexity adaptation of FLAFs, we derive a class of linear-in-the-parameter nonlinear adaptive algorithms in the frequency-domain.
%
%
\section{Functional Expansions for Low-Complexity Implementations}
\label{sec:feb}
One of the most crucial aspects in the FLAF model described in Section \ref{sec:flaf} is represented by the choice of the functional link set ${\boldsymbol{\Phi}}$, which strongly depends on the signal nature and the kind of application considered. Now, we try to understand how to choose the best possible expansion for online applications, like NAEC. In particular, we first evaluate the functional expansion types and then analyze their computational complexity.

\subsection{Functional Expansion Types}

\subsubsection{Chebyshev Functional Links}
The main feature of Chebyshev polynomial functions lies in their powerful nonlinear approximation capability, which make them useful in many learning models, from adaptive filters to artificial neural networks \cite{patra2002nonlinear, carini2016study, zhao2008functional, ZhaoTSMC2010, ComminielloWIRN2017}. 
The Chebyshev functional links are obtained by a power series expansion, including functions of previously computed functions, which leads to a very good functional approximation in proximity of the expanded sample. On the other hand, the error may increase for high-order expansions. Chebyshev functional links are quite computationally efficient and cheap compared to other power series for low-order expansions.

Consider the $i$-th input sample of the nonlinear input vector $\mathbf{x}_{\rFL,n}$, $i=0,\ldots,M_{\ri}$, the Chebyshev functional link expansion can be expressed as:
\begin{align}  \label{equ7}
	{\varphi}_j(x[n-i])  &= 2x[n-i]{\varphi}_{j-1}(x[n-i]) \nonumber \\
&	-{\varphi}_{j-2}(x[n-i]) 
\end{align}

\noindent for $j=0,\ldots,P-1$, where $P$ represents the expansion order. The overall number of functional links in a Chebyshev set is equal to the expansion order, i.e., $Q=P$. Chebyshev functional links can be initialized as follows (i.e., for $j = 0$):
\begin{align}  \label{equ8}
	&{\varphi}_{-1}(x[n-i])  = x[n-i] \nonumber \\
	&	{\varphi}_{-2}(x[n-i]) =1.
\end{align}

\subsubsection{Legendre Functional Links}
Legendre functional links are very similar to Chebyshev functions, as they also derive from a power series expansion. Such functions have been widely used for several kinds of application \cite{gil2013nonlinear, patra2008legendre}. Their popularity is due to their ability to arbitrarily well approximate any finite-memory continuous nonlinear system \cite{CariniSIGPRO2015}.

The Legendre functional link expansion can be expressed for the $i$-th input sample as:
\begin{align}  \label{equ9}
	{\varphi}_j(x[n-i])  &= \frac{1}{j}\{(2j-1)x[n-i]{\varphi}_{j-1}(x[n-i]) \nonumber \\
	&	-(j-1){\varphi}_{j-2}(x[n-i]) \}
\end{align}

\noindent with $ j = 0,\ldots, P -1$. Similarly to the Chebyshev functional link expansion, the overall number of functional links in the Legendre set is $Q = P$. The initialization is also equal to the Chebyshev one of \eqref{equ8}.

\subsubsection{Trigonometric Functional Links}
Compared to other expansions, the trigonometric functional links lead to a very compact representation in the mean square sense of every single nonlinear function \cite{pao1989adaptive, PatraTSMC1999}. Trigonometric functional links show less computational complexity and better performance than other power-series-based nonlinear expansions \cite{ComminielloTASL2013, SicuranzaTASL2011}. This is the reason why they have been widely adopted in several signal processing applications \cite{ComminielloTASL2013, SicuranzaTASL2011, PatelTCASI2016, CariniICASSP2017, ZhangTNNLS2018}.

The trigonometric functional link expansion is defined for the $i$-th entry as:
\begin{align}  \label{equ10}
	\varphi_j(x[n-i])  =
	\begin{cases}
	\sin ( p \pi x[n-i]),~~~~ j=2p-2 \\
	\cos ( p \pi x[n-i]),~~~~ j=2p-1
	\end{cases}
\end{align}

\noindent where $j = 0,\ldots, Q-1$ and $p = 1,\ldots, P$. Unlike the previous two cases, the overall number of functional links here is $Q = 2P$. It is worth noting that eq.~\eqref{equ10} refers to a memoryless expansion (further details in \cite{ComminielloTASL2013}).

\subsubsection{Random Vector Functional Links}
As described in \cite{pao1994learning} and \cite{igelnik1995stochastic}, the random vector (RV) functional link expansion is parametric with respect to a weight matrix, whose values are randomly selected \cite{igelnik1995stochastic, ScardapaneIS2016, ZhangTCYB2017, XuTSMC2019}. As a matter of fact, the input vector at the $n$-th time instant $ \mathbf{x}_{\rFL,n}$ is subject to a randomization step that returns the vector $ \mathbf{z}_n \in \mathbb{R}^{M_{\re}}$:
\begin{align}  \label{equ11}
	\mathbf{z}_n=\mathbf{V}\mathbf{x}_{\rFL,n}+ \mathbf{b}
\end{align}

\noindent where $\mathbf{V}\in \mathbb{R}^{({M_{\re} \times M_{\ri}})} $ and $ \mathbf{b}\in \mathbb{R}^{M_{\re}}$ are randomly drawn from a uniform probability distribution in $[-1,+1]$. The resulting vector $ \mathbf{z}_n$ is then further processed by using a sigmoid function:
\begin{align}  \label{equ12}
	\varphi(z[n-i])  = \frac{1}{1+e^{(-z[n-i])}},
\end{align}

\noindent which represents the only nonlinearity involved in the expansion process (i.e., $Q = 1$). Such a function is applied to every sample of the expanded vector $ \mathbf{z}_n$. It is worth noting that the expansion length $M_{\re}$ in this case does not depend on any expansion order but it can be set \textit{a priori}. Moreover, any other nonlinear function can be chosen instead of the sigmoid in \eqref{equ12}.

\subsubsection{Adaptive Exponential Functional Links}
Adaptive exponential (AE) functional links are obtained by taking the trigonometric series expansion for the $i$-th input sample and by multiplying each function by an exponential term, whose argument contains an adaptive parameter \cite{PatelTCASI2016}:
\begin{align}  \label{equ99}
	\varphi_j(x[n-i])  =
	\begin{cases}
	e^{-a\left[n\right] \left|x\left[n-i\right]\right|} \sin \left( p \pi x\left[n-i\right]\right), j=2p-2 \\
	e^{-a\left[n\right] \left|x\left[n-i\right]\right|} \cos \left( p \pi x\left[n-i\right]\right), j=2p-1
	\end{cases}
\end{align}

\noindent where $P$ is the order of functional expansion and $a\left[n\right]$ is the adaptive exponential factor that is updated by using a gradient descent rule \cite{PatelTCASI2016}.

The AE expansion improves the performance of the trigonometric functional links, as the adaptive exponential term is able to overcome any possible amplitude limitation of the trigonometric functions.
%
%
%
%
%
\subsection{Computational Analysis of the Functional Expansions}
Now we compare the types of functional links described above in terms of the computational resources required by the expansions.

The Chebyshev functional link expansion in \eqref{equ7} involves $ 2 M_{\re} = 2PM_{\ri} $ multiplications and $ M_{\re} = PM_{\ri} $ additions at each iteration. The Legendre functional link expansion in \eqref{equ9} is very similar to the Chebyshev one, but it requires slightly larger resources. In fact, the Legendre expansion involves $ 4M_{\re} = 2P(2M_{\ri} + 1) $ multiplications and $ P(M_{\ri} + 2) $ additions. Memoryless trigonometric function links of \eqref{equ10} has been largely investigated in the literature (e.g., \cite{ComminielloTASL2013}). Memoryless trigonometric functional links show $ M_{\re}/2 + P = P(M_{\ri} + 1) $ multiplications and $ M_{\re} = 2PM_{\ri} $ function evaluations (e.g., sines and cosines, which can be easily implemented by lookup tables). RV functional links in \eqref{equ11} and \eqref{equ12} involve $ M_{\re} (M_{\ri} + 1) $ multiplications, $ M_{\re} (M_{\ri} + 1) $ additions and $ M_{\re} $ function evaluations for each iteration. The addition cost required from the initialization of $\mathbf{V}$ and $\mathbf{b}$ is negligible with respect to the rest of the process. Moreover, since $M_{\re} \geq M_{\ri}$, the RV expansion has a computational cost that is approximately proportional to the quadratic order with respect to the length of the nonlinear output vector. Finally, the AE expansion is very similar to the trigonometric one, but it also requires an additional load due to the exponentiation (i.e., $3M_{\re}$ multiplications and $M_{\re}$ function evaluations) and the adaptation of the exponential factor (i.e., $2M_{\re} + 1$ multiplications and $2M_{\re} + 2$ additions). The overall computational cost in terms of multiplications is of the order of $11M_{\re}/2 + P + 1 = 11PM_{\ri} + P + 1$.
\begin{table}[t]
	\centering
	\caption{Computational cost comparison of different functional link expansions in terms of multiplications.}  
	\begin{tabular}{l c c c c c} 
		\toprule 
		\cmidrule(l){2-2} 
		\textbf{Expansion Type} &  \textbf{No. Multiplications}\\ 
		\midrule 
		Chebyshev FL Expansion & $2PM_{\ri}$  \\ 
		Legendre FL Expansion &  $2P\left(2M_{\ri} + 1\right)$ \\ 
		Trigonometric FL Expansion & $ P\left(M_{\ri} + 1\right) $  \\ 
		Random Vector FL Expansion & $M_{\re}\left(M_{\ri}+1\right)$  \\ 
		Adaptive Exponential FL Expansion & $11P{M_{\ri}}+P+1$  \\ 
			\midrule
	\end{tabular}
		\label{tab:expansion} 
\end{table}

For a fair comparison of the functional link expansions, we may fix the expanded vector length $M_{\re}$, which implies a different number of input samples for each expansion. Alternatively, we may fix the number of input samples $M_{\ri}$ for each iteration, which produces a different number of nonlinear coefficients for each expansion. The comparison in the latter case is shown in Table \ref{tab:expansion} in terms of multiplications only. Taking into account that $ P \ll M_{\ri}$, we can say that the trigonometric functional link expansion involves the lowest complexity in terms of multiplication number, while the highest computational complexity is shown by the RV expansion. It is also worth noting that AE FLAFs, while providing very interesting performance, cannot be considered for low-cost online applications. Chebyshev functional links, instead, can be considered for low-complexity FLAF implementations if the expansion order is chosen quite small. 
%
%
%
%
\section{The Proposed Family of FLAFs in the Frequency Domain}
\label{sec:fdflaf}
We have seen in the previous section how to choose a suitable type of functional link expansion that requires limited computational resources. However, another part of an LIP nonlinear filter is highly computational demanding and it is represented by the parameter adaptation process. In this section, we show how to design efficient adaptation for LIP nonlinear filters and in particular for FLAFs.

With respect to other nonlinear models, the FLAF is very effective in modeling nonlinearities. However, its efficiency is not always optimized and it may represent a limitation for the use of such methods in online applications, where there is a strong need to reduce any input delay or latency or computational load. To overcome such issues and reduce the computational complexity, we propose a frequency-domain implementation of the model, including the adaptation rule of the model parameters. To this end, we aim at implementing the adaptation scheme of the FLAF in the frequency-domain, thus developing a family of frequency-domain FLAFs (FD-FLAFs). With respect to classic linear frequency-domain adaptive filters, FD-FLAFs involve an additional branch that includes the transformed nonlinear information of the input signal, which is fundamental for the nonlinear modeling. This information is additional with respect to the linear contribution processed by the linear frequency-domain algorithms. The analysis in the Fourier domain of such additional information is different from the linear contribution as it is prevalently related to the distortions affecting the input signal. Indeed, the filter on the nonlinear branch models the nonlinearities rather than modeling the acoustic path as it happens in the linear branch and in the classic frequency-domain adaptive filters. We consider the linear and the nonlinear contribution in a joint adaptation that makes the algorithm formulations more compact and easier to be implemented.

In the following, we introduce the FD-FLAF in its overlap and save configuration and then we further improve the model by considering the partitioned-block implementation in the frequency-domain of the FLAF (namely the PBFD-FLAF).
%
%
%
%
\subsection{Overlap-Save Frequency-Domain FLAF}
\label{subs:osfdnew}
In this subsection, we define the FLAF model in the frequency-domain that we simply denote as FD-FLAF. This model adopts an adaptation scheme based on the overlap and save approach in the frequency domain (namely, the OS-FDAF). The OS-FDAF algorithm is the frequency-domain equivalent version of the block least-mean square algorithm \cite{ferrara1980fast,clark1983unified}. It converges in the mean to the optimum Wiener solution \cite{shynk1992frequency}. The update rule of the OS-FDAF can be seen as a transposition in the frequency-domain of the block least mean square algorithm update rule with some additional constraint \cite{shynk1992frequency, uncini2015fundamentals}, i.e.:
\begin{align}  \label{equ13}
	\mathbfit{W}_{k} = \mathbfit{W}_{k-1} + \left(\bL_k \mathbfit{X}_k\tH  \mathbfit{E}_k\right)_{G},
\end{align}

\noindent where $\mathbfit{X}_k \in \mathbb{C}^{(N+N_{\re}) \times 1}$ is the frequency-domain input frame, $\mathbfit{W}_k \in \mathbb{C}^{(N+N_{\re}) \times 1}$ is the filter vector in the frequency domain, $(\cdot)\tH$ is the Hermitian operator. The notation $\left(\cdot\right)_G$ represents the windowing or gradient constraint, which is necessary to avoid any aliasing phenomena in the gradient calculation. In the frequency-domain process, this constraint is implemented within the fast Fourier transform (FFT) computation. It is worth noting that the adaptation equation \eqref{equ13} involves the joint input and weight vectors in the frequency domain, respectively as \eqref{equ5} and \eqref{equ6}. 

In \eqref{equ13}, the matrix $\boldsymbol{\Lambda}_k$ is a diagonal matrix containing the step sizes for each frequency bin, i.e., $\bL_k = \diag\left(\left[ {{\begin{array}{*{20}c}{\boldsymbol{\mu}_{\rL,k}} & {\boldsymbol{\mu}_{\rFL,k}} \\ \end{array}}}\right]\right) = \diag\left(\left[ {{\begin{array}{*{20}c} {\mu_{k}\left(0\right)} & {\mu_{k}\left(1\right)}\\ \end{array}}}\right.\right.$ $\left.\left. {{\begin{array}{*{20}c} \ldots  & {\mu_{k}\left(N+N_{\re}-1\right)} \\ \end{array}}}\right]\right)$. Thus, in the frequency domain, the convergence of one filtering branch is independent of the other one. 

\begin{algorithm}[t]
	\noindent\caption{{\textbf{Summary of the FD-FLAF.}}}
	\begin{algorithmic}
			\item[] Init.: $\mathbfit{W}_0 = \mathbf 0, B_0\left(m\right) =\delta_m ~\text{for}~ m =0,1,\ldots, N+N_{\re}-1$
		\item[] \textbf{for} $ k = 0,1, \ldots $, and for each block of $ L$ samples
		\item[] ~~~~$ \mathbf{x}_{\rL,k} \leftarrow \left[\mathbf{x}_{k}^{(M)} ~~\mathbf{x}_{\text{new}}^{(L)}\right] $ and $ \mathbf{g}_k \leftarrow \left[\mathbf{g}_{\text{old}}^{(M_{\re})} ~~\mathbf{g}_{k}^{(L_{\re})}\right] $
		\item[] ~~~~$ \mathbf{x}_{k} \leftarrow \left[\mathbf{x}_{\rL,k} ~~\mathbf{g}_k\right] $ 
	\item[] ~~~~$ \mathbfit{X}_{k} =  \text{FFT}\left(\mathbf{x}_k\right) $ 
	\item[] ~~~~$ \mathbf{y}_{k} = \text{IFFT}(\mathbfit{X}_{k} \mathbfit{W}_{k})^{\lfloor L+L_{\re} \rfloor} $ 
	\item[] ~~~~$ \mathbfit{E}_{k}=\text{FFT}\left(\left[\mathbf 0~~ {\mathbf{d}}_k - {\mathbf{y}}_{k} \right]\right) $ 
	\item[] ~~~~$  B_{k}\left(m\right)=\lambda B_{k-1}\left(m\right)+\left(1-\lambda\right)\left|\mathbfit{X}_{k}\left(m\right)\right|^2~ \forall m$
	\item[] ~~~~$\bL_k = \diag\left(\left[{{\begin{array}{*{20}c} {\mu_{k}\left(0\right)} & \ldots & {\mu_{k}\left(N+N_{\re}-1\right)} \\ \end{array}}}\right]\right)$
	\item[] ~~~~$ \nabla J_{k} = \bL_{k} \mathbfit{X}_{k}\tH \mathbfit{E}_{k}$ 
	\item[] ~~~~$ \left(\nabla J_{k}\right)_G = \text{FFT}\left( {\left[ {{\begin{array}{*{20}c} {\text{IFFT}\left( {\nabla J_k } \right)^{\left\lceil M + M_{e} \right\rceil }} \hfill \\ {{\mathbf{ 0}}_{L+L_{re}} } \hfill \\
										 \end{array} }} \right]} \right)$ 
	\item[] ~~~~$ \mathbfit{W}_{k}=\mathbfit{W}_{k-1} + \left(\nabla J_{k}\right)_G $ 
	\item[] \textbf{end for}	
	\end{algorithmic}
\end{algorithm}
\setlength{\textfloatsep}{0.2cm}
\setlength{\floatsep}{0.2cm}
\begin{figure*}[t]
	\centering
	\includegraphics[width=0.85\textwidth,keepaspectratio]{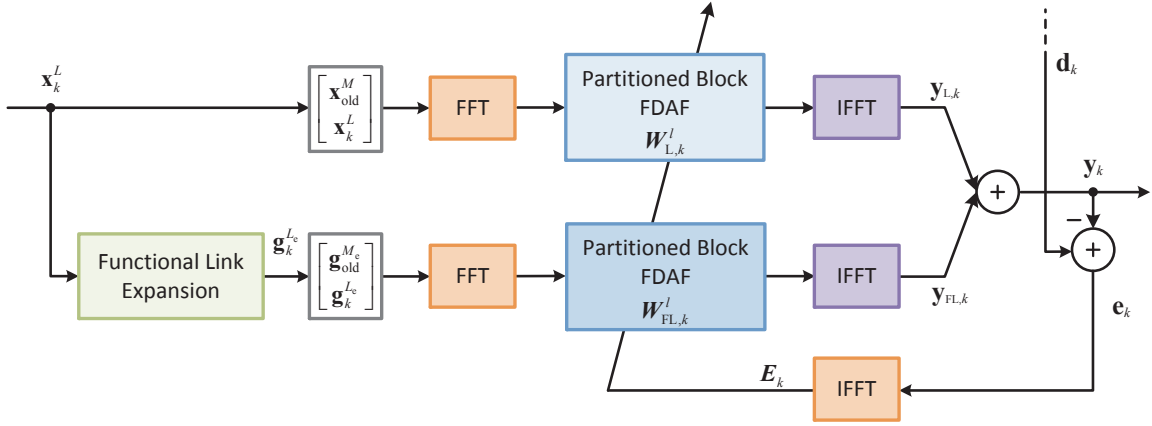}
	\caption{Scheme of the proposed Partitioned-Block Frequency-Domain FLAF algorithm.}
	\label{fig:Schematic}
\end{figure*}

In order to speed-up the slower convergence modes, we can define a power normalization rule that improves the overall convergence rate. To this end, we can derive the estimated power of the $ m $-th frequency bin $  B_{k}(m) $. Denoting with $ \mu$ a fixed step-size parameter (which can be also distinguished for the linear and the nonlinear branches), we can express the $m$-th step-size parameter of the FD-FLAF proportionally to the inverse of its power, i.e.:
\begin{equation}
    \mu_k\left(m\right) = \frac{\mu}{\left(\zeta + B_k\left(m\right)\right)},
    \label{eq:normalizedss}
\end{equation}
    
\noindent where $m = 0, \ldots, N+N_{\re}-1$, and $\zeta$ is a small positive constant avoiding divisions by zero. If the input signal is a white Gaussian noise, $  B_{k}(m) $ will be the same for each frequency bin, thus $\boldsymbol{\mu}_k = \mu \mathbf{I}$. In order to reduce the noise that could be produced by significant difference in successive step-size power values, we can compute the $m$-th power frequency bin by using a low-pass filtering \cite{narayan1981frequency}:
\begin{align} \label{equ14}
	B_{k}(m)=\lambda B_{k-1}(m)+(1-\lambda)\left|{X}_{k}(m)\right|^2,
\end{align}

\noindent for $m=0,\ldots,N+N_{\re}-1$, where $ \lambda $ is a forgetting factor and $ \left|X_{k}(m)\right|^2 $ the energy of the $ m $-th bin. The step-size normalization leads the OS-FDAF to achieve both reduced complexity and higher convergence rate with respect to the block least mean square algorithm. 

The implementation of the proposed FD-FLAF method involving the overlap and save approach is summarized in Algorithm 1. One of the drawbacks of this algorithm is represented by the introduction of a processing delay due to the FFT computation for each block of input samples. We address this limitation in the next subsection.
%
%
%
%
%
%
\subsection{Partitioned-Block Frequency-Domain FLAF}
Here we try to further improve the implementation of the frequency-domain adaptation by focusing on reducing the delay introduced by the OS-FDAF algorithm. Indeed, although such an algorithm shows a low computational load and convergence properties comparable with time-domain algorithms, the latency due to block processing still represents a challenging issue to be solved. Such delay between input and output is at least equal to the overall block length $L+L_{\re}$. 
\begin{algorithm}[t]
	\noindent\caption{{\textbf{Summary of the PBFD-FLAF.}}}
	\begin{algorithmic}
		
		\item[] Init.: $\mathbfit{W}_0  = \mathbf 0, B_0(m) =\delta_m ~\text{for}~ m =0,1,\ldots, N + N_{\re}-1$
		\item[] \textbf{for} $ k = 0,1, \ldots $, for each block of $ L+L_{\re} $ samples
        \item[] ~~~~$ \mathbf{x}_{kL}^0 \leftarrow \left[\mathbf{x}_{\text{old}}^{{0,M}} ~~\mathbf{x}_{k}^{{0,L}}\right] $ and $ \mathbf{g}_k^0 \leftarrow \left[\mathbf{g}_{\text{old}}^{(0,M_{\re})} ~~\mathbf{g}_{k}^{(0, L_{\re})}\right] $
        \item[] ~~~~$ \mathbf{x}_{k}^0 \leftarrow \left[\mathbf{x}_{kL}^0 ~~\mathbf{g}_k^0\right] $
		\item[] ~~~~$ \mathbfit{X}_{k}^0 = \text{diag} \left( \text{FFT}\left(\mathbf{x}_{k}^0\right)\right) $
		\item[] ~~~~$ {\mathbf{y}}_{k} = \text{IFFT}\left(\sum_{l=0}^{M_{P}-1} \mathbfit{X}_{k-pl}^0 \mathbfit{W}_{k}^l)\right)^{\lfloor L+L_{\re} \rfloor}$
		\item[] ~~~~$  B_{k}\left(m\right)=\lambda B_{k-1}\left(m\right)+\left(1-\lambda\right)\left|\mathbfit{X}_{k}^0\left(m\right)\right|^2~ \forall m$
		\item[] ~~~~$\bL_k = \diag\left(\left[{{\begin{array}{*{20}c} {\mu_{k}\left(0\right)} & \ldots & {\mu_{k}\left(N+N_{\re}-1\right)} \\ \end{array}}}\right]\right)$
		\item[] ~~~~~~~~\textbf{for} $ l=0,\ldots,M_P-1 $
		\item[] ~~~~~~~~~~~~$ \mathbfit{E}_{k}=\text{FFT}\left(\left[{\mathbf{0}_{M+M_{\re}}}~~ {\mathbf{d}}_k- {\mathbf{y}}_{k}\right]\right) $
		\item[] ~~~~~~~~~~~~$ \nabla J_{k}^l = \bL_{k}^l {\mathbfit{X}_{k-pl}^{0,\text{H}}} \mathbfit{E}_{k}$ 
		\item[] ~~~~~~~~~~~~$ \left(\nabla J_{k}^l\right)_G = \text{FFT}\left( {\left[ {{\begin{array}{*{20}c} {\text{IFFT}\left( {\nabla J_k }^l \right)^{\left\lceil M + M_{e} \right\rceil }} \hfill \\ {{\mathbf{ 0}}_{L+L_{re}} } \hfill \\
										 \end{array} }} \right]} \right)$
		\item[] ~~~~~~~~~~~~$ \mathbfit{W}_{k}^l=\mathbfit{W}_{k-1}^l + \left(\nabla J_{k}^l\right)_G $ 
		
		\item[] ~~~~~~~~\textbf{end for}
		\item[] \textbf{end for}

	\end{algorithmic}
\end{algorithm}
\setlength{\textfloatsep}{0.2cm}
\setlength{\floatsep}{0.2cm}

In order to solve this problem, we adopt a partitioned-block approach, which divides the filter in $M_P$ sub-filters, thus reducing the overall delay to $M_P$ samples. Therefore, $(M+M_{\re})/M_P$ number of smaller convolutions are carried out in the frequency domain, but only the first partition introduces a delay. The achieved latency reduction makes this approach suitable for real-time applications. 

We apply such an approach to the frequency-domain FLAFs, thus designing the partitioned-block frequency-domain FLAFs (PBFD-FLAFs). This algorithm is able to further improve the efficiency of this class of LIP nonlinear filters for online applications, while keeping effective performance. The implementation of the proposed PBFD-FLAF method is summarized in Algorithm 2.

\section{Performance Analysis of the PBFD-FLAF}
\label{sec:pbcomplexity}
To evaluate the performance of PBFD-FLAF, we assume that frequency bins are uncorrelated to each other \cite{uncini2015fundamentals}. Therefore, the correlation matrix of the input signal can be approximated to a diagonal matrix. However, the performance may change based on the amount of samples considered in each input block.

\subsection{Performance Analysis in the Case of $L=M$}
Let us consider the length of the input block as $L=M$. From the definition of the partitioned-block approach we have $\mathbfit{X}_k^l = \mathbfit{X}_{k-l}^0 $, thus the vector $ \mathbfit{X}_{m,k} $ containing the $m$-th frequency bin of the $k$-th block can be expressed as:
\begin{align}  \label{equ16}
	\mathbfit{X}_{m,k}&=\left[X_k^0\left(m\right)~~X_k^1\left(m\right)~\ldots~X_k^{M_P-1}\left(m\right)\right]\tT \nonumber \\ &=\left[X_k^0\left(m\right)~~X_{k-1}^0\left(m\right)~\ldots~X_{k-M_P+1}^{0}\left(m\right)\right]\tT.
\end{align}

\noindent The vector $\mathbfit{X}_{m,k}$ contains the $M_P$ samples of the input block for the $m$-th frequency bin. Thus, the convergence rate of each frequency bin relies on the eigenvalues of the $ M_P\times M_P$ correlation matrix of the input: 
\begin{align}  \label{equ17}
	\mathbf{R}_{m,k} = \text{E}\left\{\mathbfit{X}_{m,k}\mathbfit{X}_{m,k}\tH\right\},
\end{align}

\noindent which can be also written in a normalized form as:
\begin{align}  \label{equ18}
	\overline{\mathbf{R}}_{m,k} = \left(\text{diag}\left(\mathbf{R}_{m,k}\right)\right)^{-1}\mathbf{R}_{m,k}.
\end{align}

In order to compute the correlation matrix, we start considering a white noise sequence as input $ x\left[n\right] $. For, $L=M$, the discrete Fourier transform (DFT) can be computed using $ 2M $ points and it can be expressed for the $m$-th frequency bin as: 
\begin{align}  \label{equ19}
	X_k^0\left(m\right)=\sum_{n=0}^{2M-1} x\left[kM-M+n\right] e^{-j\frac{2\pi}{2M}m n}.
\end{align}

\noindent From the previous considerations, we have:
\begin{align}  \label{equ20}
	\text{E}\left\{X_k^0\left(m\right)X_{k-1}^{0*}\left(m\right)\right\} = 
	\begin{cases}
		2M\sigma_x^2~~~ &\text{for}=l=0\\
		(-1)^{m} \times M\sigma_x^2~~~ &\text{for}=\pm1\\
		0 &\text{otherwise}
	\end{cases},		
\end{align}

\noindent where $ \sigma_x^2$ is the variance of $x\left[n\right]$. Generalizing the result of \eqref{equ20}, it is possible to define the normalized correlation matrix for a white input signal as:
\begin{align} \label{equ21}
	\overline{\mathbf{R}}_{m,k} &= \left(\text{diag}\left(\mathbf{R}_{m,k}\right)\right)^{-1}\mathbf{R}_{m,k}\nonumber\\
	&  =\left[
	\begin{matrix}
		1 & \alpha_m & 0 & \ldots & \ldots &0\\
		\alpha_m & 1 & \alpha_m & 0 & \ldots &0 \\
		0 & \alpha_m & 1 & \alpha_m & \ddots &\vdots \\
		\vdots & 0 & \alpha_m & 1 & \ddots &0 \\ 
		\vdots & \vdots & \ddots & \ddots & \ddots &\alpha_m \\
		0 & 0 & \ldots & 0 & \alpha_m & 1 \\
	\end{matrix}
	\right],
\end{align}

\noindent where $ \alpha_m = (-1)^m \times 0.5 $ depends on the overlap between two successive frames. In the case of a $50\%$ overlap, we have $\alpha_m=\pm0.5$. 

The convergence properties can be evaluated by computing the eigenvalues of the normalized correlation matrix $\overline{\mathbf{R}}_{m,k} $. Indeed, from \eqref{equ20}, we can derive the condition number ${\chi}(\overline{\mathbf{R}}_{m,k})=\lambda_{max}/\lambda_{min}$, which is independent of the frequency index $m$ and increases as the number of partitions $M_P$ grows. On the other hand, the conditioning number decreases proportionally to the factor $ \vert \alpha_m\vert$. As a result, it is more convenient to implement the partitioned-block algorithm with an overlap of less than $ 50\% $, which implies $ L<M $.
  
\subsection{Performance Analysis in the Case of $ L<M $}
Let us consider now $L=M/p$, where $p$ is a positive integer. The DFT involves $(M+L)$ points and it can be defined for the $m$-th frequency bin as:
\begin{align}  \label{equ22}
	X_k^0\left(m\right)=\sum_{n=0}^{M+L-1} x\left[kL-M+n\right]e^{-j\frac{2\pi}{M+L}m n}.
\end{align}

\noindent Even in this case, for a white input sequence $x\left[n\right]$, we have:
\begin{align}  \label{equ23}
	\alpha_m=\frac{1}{p+1}e^{j(2\pi pi/p+1)},
\end{align}

\noindent which means that the parameter $\alpha_m$ tends to decrease as the overlap increases. Moreover, to avoid any convergence issue due to the aliasing of the Fourier transform, it is possible to use the filter update involving the gradient constraint, which guarantees the same convergence of a filter without partitioning.

It is worth noting that the nonlinear path of the PBFD-FLAF follows the same analysis of this latter case, as the length $L_{\re}$ is always smaller than $M_{\re}$.
%
%
%
%

\section{Experimental Results}
\label{sec:results}
In this section, the performance evaluation of the new PBFD-FLAF algorithm will be discussed. PBFD-FLAFs performance has been evaluated from different perspectives. We have considered both the computational complexity and the overall best setup of the filters. In all the experiments, we have evaluated the performance contribution given by the five types of the functional link expansions, i.e., the Chebyshev, the Legendre, the trigonometric, the random vector and adaptive exponential expansions. Experiments have been conducted in MATLAB.

Various simulation scenarios have been considered, related to teleconferencing environments with different reverberation times. The acoustic environments are characterized by simulated impulse responses sampled at 8 kHz. We have used different far-end input signals, including both colored noise signals and real speech and audio signals. The colored input noise is obtained by applying a first-order autoregressive model to a white Gaussian noise, according to the following function: $ \sqrt{1- \alpha^2} / (1-\alpha z^{-1})$ , with $\alpha = 0.8$. For each experiment, the desired signal $d\left[n\right]$, representing the near-end microphone signal, includes the contribution of the reverberated far-end signal with any additive white Gaussian noise with a specific signal-to-noise ratio (SNR).   

We have utilized the symmetrical soft-clipping nonlinearity to assess the performance of the proposed algorithms. The unknown system to be identified is composed of a cascade of a nonlinear block followed by a linear block. The nonlinear subsystem applies a soft clipping nonlinearity to the input signal to simulate the classic saturation impact of a loudspeaker, as described in \cite{comminiello2014nonlinear}. The symmetrical soft-clipping can be written as:
\begin{align}  \label{soft-clipping}
	\overline y\left[n\right]= \left\{ \begin{array}{ccl}
		2x\left[n\right]/3 \zeta & \mbox{for} & 0\leq x\left[n\right]\leq\zeta \\[0.5ex]
		 \text{sign} \left(x(n) \right)\frac{3-(2-[x]|/ \zeta)^2}{3} & \mbox{for} & \zeta\leq x\left[n\right]\leq 2 \zeta \\[0.5ex]
		\text{sign} \left(x(n)\right) & \mbox{for} & 2\zeta\leq x\left[n\right]\leq 1
	\end{array}\right.
\end{align} 

\noindent where $ 0<\zeta \leq 0.5$ is a nonlinearity threshold. The following linear subsystem system is described by the input-output relationship:
\begin{equation}  
    \begin{split}\label{equ27}
	\overline{y}\left[n\right] &= \frac{6}{10}\sin^3\left[\pi x\left[n\right]- \frac{2}{x^3\left[n\right]+2}\right]\\
	&- \frac{1}{10}\cos\left[4 \pi x\left[n-4\right]\right]+1.125.
	\end{split}
\end{equation}


Performance is evaluated based on different quality measures. The most significant and popular quality index that can be considered in NAEC applications is the echo return loss enhancement (ERLE), which indicates the amount of echo signal that is canceled by the algorithm. The ERLE is defined as:  
\begin{equation}
	{\rm{ERLE}}\left[n\right]= 10\log_{10}\left(\frac{\E\left\{d^2\left[n\right]\right\}}{\E\left\{e^2\left[n\right]\right\}}\right),
	\label{equ24}
\end{equation}

\noindent where $\E \{\cdot\}$ represents the expectation operator. 
\begin{table}[t]
	\centering 
	\caption{Performance comparison in terms of objective measures between Linear-PBFDAF and PBFD-FLAF with all types of expansion in case of colored noise input affected by a symmetrical soft-clipping nonlinearity.}
	\begin{tabular}{l c c c c c} 
		\toprule 
		& \multicolumn{2}{c}{\textbf{Measure}} \\ 
		\cmidrule(l){2-3} 
		\textbf{Filter} & ERLE Mean &  TIME (Sec.)\\ 
		\midrule 
		Linear PBFDAF & 8.44 &  0.13  \\ 
		PBFD-FLAF Tri & 12.24 &  0.93  \\ 
		PBFD-FLAF Che & 12.25 &  0.64 \\ 
		PBFD-FLAF Leg & 12.26 &  0.64 \\ 
		PBFD-FLAF RV & 10.76 &  0.60  \\ 
		PBFD-FLAF AE & 12.17 &  0.92  \\ 
		\midrule
	\end{tabular}
		\label{tab:colored1} 
\end{table}
\begin{figure}[t]
	\centering
	\includegraphics[width=0.49\textwidth,keepaspectratio]{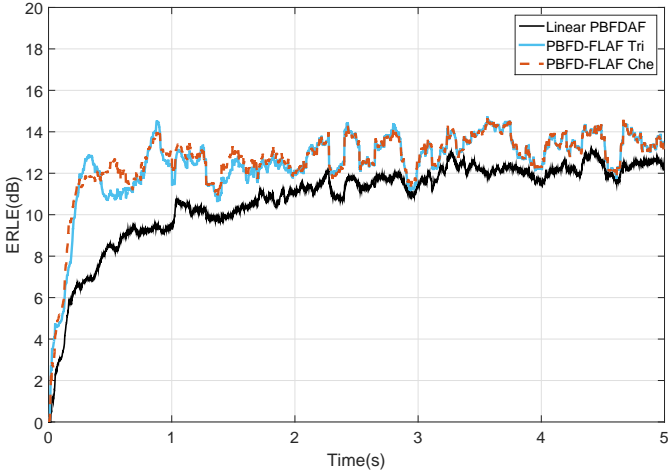}
	\caption{Performance comparison in terms of ERLE between Linear-PBFDAF and PBFD-FLAF with all types of expansion in case of colored noise input affected by a symmetrical soft-clipping nonlinearity.}
	\label{fig:colored1}
\end{figure}

However, the ERLE does not completely show the real increment of both quality and intelligibility of the signal. Thus, together with the ERLE, we can also use other measures that are appropriately designed for speech and audio signals. In that sense, one of the most fundamental metrics for the quality assessment of a signal is the perceptual evaluation of speech quality (PESQ) objective measure, which determines the quality of speech by approximating the overall loudness difference between the original signal and its approximation. 
The original clean signal and its estimated version, respectively $s \left[n\right]$ and $y \left[n\right]$, are equalized to a standard listening level and then filtered considering a response approximating a standard telephone handset. Such signals are transformed and represented in terms of loudness spectra. The loudness difference between the $s \left[n\right]$ and $y \left[n\right]$ is averaged over time and frequency to generate a subjective quality score between 1.0 and 4.5, where the higher the value, the better the quality. Another objective speech-intelligibility measure that is suitable for this kind of evaluation is the short-time objective intelligibility (STOI), whose output is expected to have a monotonic relation with the subjective speech intelligibility. A higher value of the STOI corresponds to a high intelligibility of the speech signal. 

\subsection{Experiment Set 1: Stationary Conditions}
\label{subs:Experi1}
In general, it is not trivial to investigate the performance of the proposed PBFD-FLAFs and compare them to the linear PBFDAF. The comparison is done in a situation that is as fair as possible, which means that the buffer lengths, the number of FFT coefficients, and the rest of the parameter setup must be the same for all the filters. We consider the system described by \eqref{soft-clipping} and we use the same filtering rules for both the PBFDAF and the PBFD-FLAFs. We consider a simulated acoustic impulse response with a reverberation time of $T_{60} = 150$ ms and truncated to 320 samples. The SNR is set to $20$ dB. The length of the experiment for real input is $10 $ seconds and for colored input is $ 5 $ seconds. 
The following parameter values summarizes the PBFD-FLAFs setup: $\zeta= 0.2 $, $M=300$, $ \mu_{\text{PB}}=0.005$, the step-size values $\mu_{\rL}=0.01$ and $\mu_{\rFL}=0.001$ for all the PBFD-FLAFs, $\delta = 10^{-3} $ for all the filters, $P=10 $, $M_{\ri} =128 $, $N_{{\rm{fft}}}= M_{\ri}$, and $M_{\text{P}} = 4$ partitions. 
\begin{table}[t]
		\centering 
		\caption{Performance comparison in terms of ERLE between Linear-PBFDAF and PBFD-FLAF with all types of expansions in case of female speech input affected by a symmetrical soft-clipping nonlinearity.}
	\begin{tabular}{l c c c c} 
		\toprule 
		& \multicolumn{3}{c}{\textbf{Measure}} \\ 
		\cmidrule(l){2-4} 
		\textbf{Filter} & ERLE Mean & STOI & PESQ \\ 
		\midrule 
        Linear PBFDAF & 5.91 & 0.928 & 2.87 \\ 
		PBFD-FLAF Tri & 13.54 & 0.938 & 3.01 \\ 
		PBFD-FLAF Che & \textbf{14.24} & \textbf{0.946} & \textbf{3.07} \\ 
		PBFD-FLAF Leg & 13.69 & 0.937 & 3.00 \\ 
		PBFD-FLAF RV & 13.71 & 0.937 & 3.00 \\ 
		PBFD-FLAF AE & 13.54 & 0.938 & 3.01 \\ 
		\midrule
	\end{tabular}
	\label{tab:real1} 
\end{table}
\begin{figure}[t]
	\centering
	\includegraphics[width=0.49\textwidth,keepaspectratio]{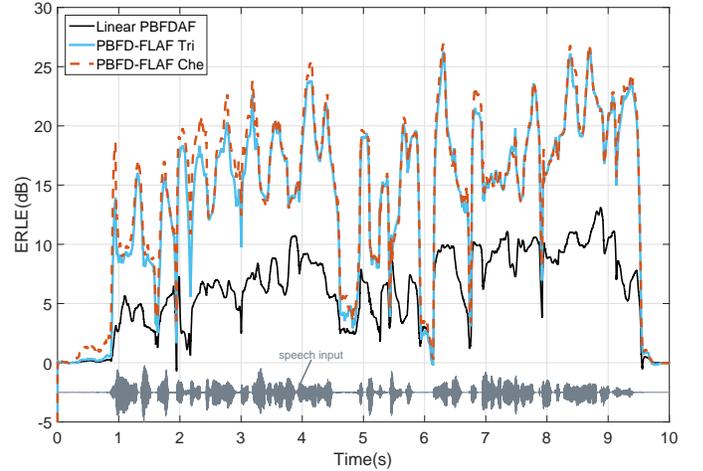}
	\caption{Performance comparison in terms of ERLE between Linear-PBFDAF and PBFD-FLAF with all types of expansions in case of female speech input affected by a symmetrical soft-clipping nonlinearity.}
	\label{fig:real1}
\end{figure}

Results are shown in terms of the online and mean ERLE, PESQ, STOI and processing time, for colored and speech input signals. We first consider the behavior of the PBFD-FLAs in the presence of colored noise input for $M_{\ri} = 128$, thus having a large number of nonlinear elements for all the models. In this situation, all the PBFD-FLAFs achieve similar performance as can be seen from Table~\ref{tab:colored1}. However, the most stable and reliable performance, even considering a smaller $M_{\ri}$, are the PBFD-FLAFs with trigonometric, Legendre and Chebyshev expansions. The ERLE behaviors of trigonometric and Chebyshev PBFD-FLAFs in time are depicted in Fig.~\ref{fig:colored1}. Most of the difference between all the PBFD-FLAFs is restricted in the convergence state, i.e., in the first 2 seconds of the experiment, after which all the filters show the same steady-state performance.
\begin{table}[t]
	\centering 
	\caption{Performance comparison in terms of objective measures between Linear PBFDAF and PBFD-FLAFs in case of colored noise input affected by a symmetrical soft-clipping nonlinearity with four different threshold values.} 
	\begin{tabular}{l c c c c c} 
		\toprule 
		& \multicolumn{2}{c}{\textbf{Measure}} \\ 
		\cmidrule(l){2-3} 
		\textbf{Filter} & ERLE Mean & TIME (Sec.)\\ 
		\midrule 
		Linear PBFDAF & 7.91 &  0.22  \\ 
		PBFD-FLAF Tri & 13.82 &  1.83  \\ 
		PBFD-FLAF Che & 13.79 &  1.33 \\ 
		PBFD-FLAF Leg & 13.83 &  1.33 \\ 
		PBFD-FLAF RV & 13.69 &  1.16  \\ 
		PBFD-FLAF AE & 13.73 &  1.70  \\ 
		\midrule
	\end{tabular}
		\label{tab:coloredth4} 
\end{table}
\begin{figure}[t]
	\centering
	\includegraphics[width=0.49\textwidth,keepaspectratio]{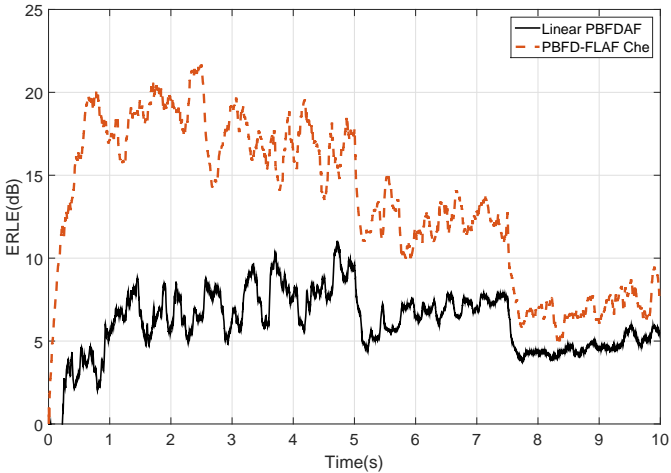}
	\caption{Performance comparison in terms of ERLE between Linear PBFDAF and PBFD-FLAFs with all types of expansions in case of colored noise input affected by a symmetrical soft-clipping nonlinearity with four different nonlinear threshold values.}
	\label{fig:coloredth4}
\end{figure}

We also try to evaluate the performance of the filters in the presence of a distorted speech signal. If we consider a high enough input buffer length, we achieve similar results for all the filters, especially those with trigonometric, Legendre and Chebyshev expansions. However, if we consider a smaller input buffer length, e.g., $M_{\ri} = 16$, which corresponds to a smaller number of nonlinear filter elements to be updated, the best results are provided by the Chebyshev PDFD-FLAF, which is the most reliable and stable solution, as it can be seen from Table \ref{tab:real1} and Fig.~\ref{fig:real1}.

\subsection{Experiment Set 2: Nonstationary Conditions with Different Nonlinearity Degree}
\label{subs:Experi2}
In this set of experiments, we assess the proposed algorithms in different nonlinear conditions and we further evaluate their tracking abilities. We consider a system with the input-output relationship given by (\ref{soft-clipping}) which is a symmetrical soft-clipping nonlinearity, we chose four different values for the nonlinearity threshold in order to have different behaviors of the functional links \cite {comminiello2017combined}. We set the nonlinearity threshold to $ \zeta = \left\{0.4,~ 0.30,~0.18,~0.08\right\} $, respectively, i.e., from slight to strong nonlinear distortion. The affected signal by nonlinearities is convolved with a simulated acoustic impulse response with $T_{60} \approx 100$ ms sampled at $8$ kHz, whose length is truncated to 512 samples. We consider an SNR equals to $20$ dB. The length of the experiment is $ 10 $ seconds, corresponding to an input signal length of $80000$ samples. The parameters of this setup are taken from the Experiment Set 1.
%
\begin{table}[t]
	\centering 
	\caption{Performance comparison in terms of objective measures between Linear PBFDAF and PBFD-FLAFs in case of female speech input affected by a symmetrical soft-clipping nonlinearity with four different threshold values.}
	\begin{tabular}{l c c c} 
		\toprule 
		& \multicolumn{3}{c}{\textbf{Measure}} \\ 
		\cmidrule(l){2-4} 
		\textbf{Filter}&ERLE Mean&STOI&PESQ\\ 
		\midrule 
		Linear PBFDAF & 3.52 & 0.897 & 2.82\\ 
		PBFD-FLAF Tri & 8.19 & 0.907 & 3.55\\ 
		PBFD-FLAF Che & \textbf{9.13} & \textbf{0.911} & \textbf{3.72}\\ 
		PBFD-FLAF Leg & 8.40 & 0.908 & 3.53\\ 
		PBFD-FLAF RV & 7.05 & 0.908 & 3.28\\ 
	    PBFD-FLAF AE & 7.22 & 0.907 & 3.61\\ 
		\midrule 
	\end{tabular}
		\label{tab:realth4} 
\end{table}
\begin{figure}[t]
	\centering
	\includegraphics[width=0.49\textwidth,keepaspectratio]{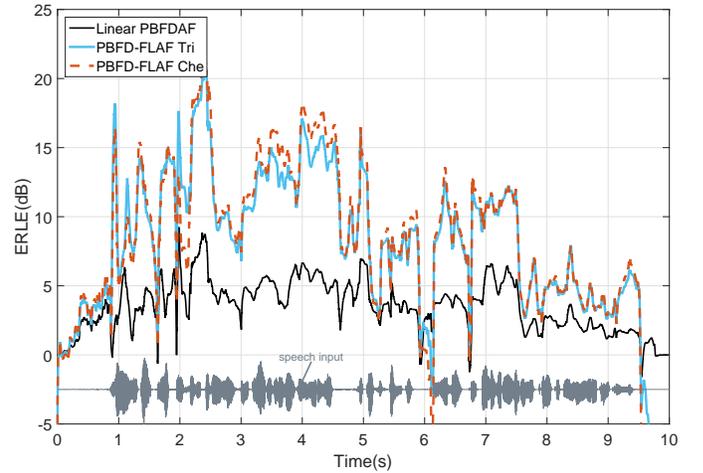}
	\caption{Performance comparison in terms of ERLE between Linear PBFDAF and PBFD-FLAFs with all types of expansions in case of female speech input affected by a symmetrical soft-clipping nonlinearity with four different nonlinear threshold values.}
	\label{fig:realth4}
\end{figure}

In the case of colored noise input, we consider a shorter input buffer length with respect to the previous experiment equal to $M_{\ri} = 32$. Again, all the PBFD-FLAFs achieve similar performance as can be seen from Table~\ref{tab:coloredth4}, with a preference for the trigonometric, Chebyshev and Legendre PBFD-FLAFs. The time behavior of the ERLEs is depicted in Fig.~\ref{fig:coloredth4}, where it is possible to notice the benefit of the PBFD-FLAFs on the linear model, even using a limited number of parameters.

In the presence of a speech signal, again the Chebyshev PBFD-FLAF slightly outperform the other filters, as shown in Table \ref{tab:realth4} and Figure \ref{fig:realth4}, appearing always rather stable, even decreasing the input buffer length, and thus the overall computational complexity.

\subsection{Experiment Set 3: Nonstationary Conditions with Different Volume Levels}
\label{subs:Experi3}
We now consider the same system identification scheme of the previous experiments but using a symmetrical soft-clipping threshold of $ \zeta = 0.21 $. While keeping a fixed nonlinearity degree, we introduce some variance in the experiment by considering different volume levels for the input signals. Signals are convolved with an impulse response of an environment with a reverberation time of $T_{60} =100ms$, sampled at $8$ kHz and truncated at $M = 320$ samples. An additional noise providing $20$ dB of SNR is considered for all the experiments. The length of each experiment is 10 seconds. We compare the PBFD-FLAFs also with the corresponding time-domain FLAFs and to a second-order Volterra filter implemented with a partitioned-block frequency-domain scheme, denoted as PBFD-VF. The comparison is fairly performed by considering the same input buffer lengths, expansion order, and parameters setup. We try to reduce as much as possible the computational complexity by setting a small input buffer length of $M_{ \rm i} =32$.

\begin{table}[t]
	\centering 
	\caption{Performance comparison in terms of objective measures between PBFD-FLAFs in case of female speech input affected by a symmetrical soft-clipping nonlinearity.}
	\begin{tabular}{l c c c c c} 
		\toprule 
		& \multicolumn{3}{c}{\textbf{Measure}}  \\ 
		\cmidrule(l){2-4} 
		\textbf{Filter} & ERLE Mean & STOI & PESQ \\ 
		\midrule 
		Linear PBFDAF & 5.35 & 0.845 & 2.85\\ 
		PBFD-FLAF Tri & 10.10 & 0.937 & 3.61\\ 
		PBFD-FLAF Che & \textbf{10.24} & \textbf{0.945} & \textbf{3.72}\\ 
		PBFD-FLAF Leg & 9.79 & 0.942 & 3.66\\ 
		PBFD-FLAF RV & 9.59 & 0.940 & 3.57\\ 
	    PBFD-FLAF AE & 9.79 & 0.942 & 3.65\\ 
		\midrule 
	\end{tabular}
		\label{tab:NLMS1} 
\end{table}
\begin{figure}[t]
	\centering
	\includegraphics[width=0.49\textwidth,keepaspectratio]{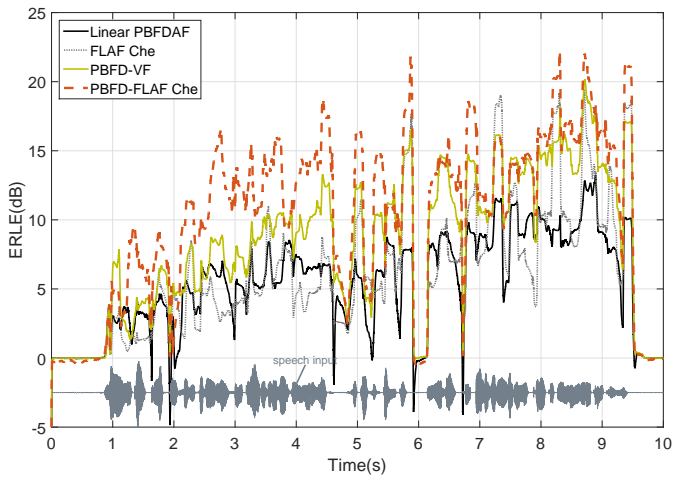}
	\caption{Performance comparison in terms of the ERLE between PBFD-FLAFs in case of female speech input affected by a symmetrical soft-clipping nonlinearity.}
	\label{fig:NLMS1}
\end{figure}

We first consider the case of a speech input signal affected by a symmetrical soft-clipping distortion. Table~\ref{tab:NLMS1} shows the comparison results in terms of mean ERLE, STOI and PESQ, and again we can see that the best performance, albeit slight in some cases, is achieved by the Chebyshev PBFD-FLAF. Figure~\ref{fig:NLMS1} shows the instantaneous ERLE comparison, by considering the best performing PBFD-FLAF, i.e., the one with the Chebyshev expansion, the linear PBFDAF, the time-domain Chebyshev FLAF, and the PBFD-VF. We can see that from the time domain Chebyshev FLAF with the same number of parameters is not able to achieve the same performance of the PBFD-FLAF. This one is also superior to the PBFD-VF, as also happens in the time domain comparisons (see for example \cite{ComminielloTASL2013}). This result shows that if we want to reduce the computational complexity of an LIP nonlinear filter, the Chebyshev expansion together with the partitioned-block implementation in the frequency domain provides the best solution.

\begin{figure}[t]
	\centering
	\includegraphics[width=0.49\textwidth,keepaspectratio]{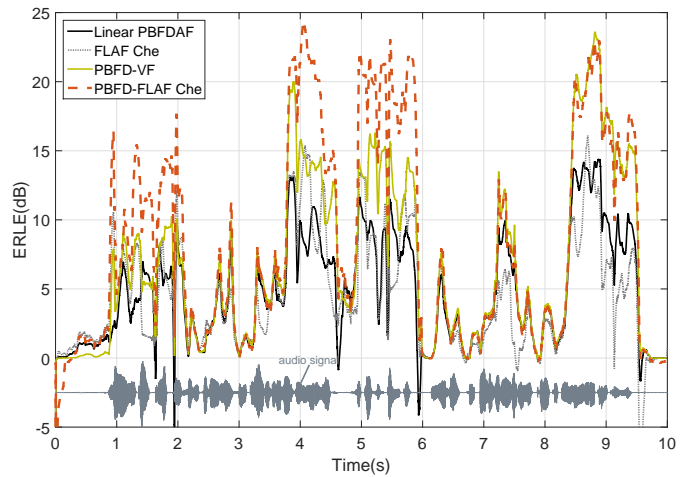}
	\caption{Performance comparison in terms of ERLE between PBFD-FLAFs in case of an audio signal input recorded from a radio station.}
	\label{fig:NLMS2}
\end{figure}

We also consider an audio input signal recorded by a radio station and capture from the reference microphone after the convolution with the impulse response. If we look at the results shown in Table~\ref{tab:NLMS2} and Fig.~\ref{fig:NLMS2}, we draw the same conclusion of the previous experiment.

\begin{table}[t]
	\centering 
	\caption{Performance comparison in terms of objective measures between PBFD-FLAFs in case of an audio signal recorded from a radio station.}
	\begin{tabular}{l c c c} 
		\toprule 
		& \multicolumn{3}{c}{\textbf{Measure}} \\ 
		\cmidrule(l){2-4} 
		\textbf{Filter}& ERLE Mean & STOI & PESQ\\ 
		\midrule 
		Linear PBFDAF & 4.73 & 0.845 & 2.85\\ 
		PBFD-FLAF Tri & 8.25 & 0.937 & 3.61\\ 
		PBFD-FLAF Che & \textbf{8.48} & \textbf{0.945} & \textbf{3.72}\\ 
		PBFD-FLAF Leg & 8.46 & 0.942 & 3.66\\ 
		PBFD-FLAF RV & 6.78 & 0.940 & 3.57\\ 
	    PBFD-FLAF AE & 8.46 & 0.942 & 3.65\\ 
		\midrule 
	\end{tabular}
		\label{tab:NLMS2} 
\end{table}

\subsection{Final Remarks on Low-Complexity Functional Links}
It is worth noting that functional link expansions mainly derive from trigonometric or polynomial series expansions that satisfy universal approximation constraints. Ideally, all the expansions are capable of modeling any nonlinearity. However, in real applications we cannot use infinite memory lengths. Moreover, in real nonlinear filtering models we should also involve some pruning method to avoid uninformative parameters that may degrade the performance. Nonlinear expansions based on trigonometric series are very powerful in approximating distorted waveform signals like speech, but a sufficiently large number of functional links are required to achieve a good approximation. This implies the necessary availability of sufficient computational resources. However, when a small set of functional links can be considered due to limited resources, polynomial series expansions, like the Chebyshev expansion, are able to model the most significant part of the nonlinearity with a very small number of parameters, as experimental results have also shown. Therefore, in conclusion, the choice of the functional link expansion strongly depends on the available resources: with large resources any expansion can provide impressive results, with a sufficient number of resources trigonometric-like functional links are the best choice, while Chebyshev functional links can be chosen for low-complexity implementation.

Concerning the open challenges, while on the one hand there is a lot of evidence of implementation of frequency-domain filters, on the other hand it will definitely be more interesting and demanding to develop efficient implementation approaches of functional expansions on different hardware devices.
%
%
%
%
%
\section{Conclusion}
\label{sec:conc}
In this paper, we have proposed a low-complexity LIP nonlinear adaptive filter for online applications that need to work with limited computational resources. To this end, we propose a family of frequency-domain functional link adaptive filters (FD-FLAFs), whose coefficient vector is updated in the frequency domain and requires fewer resources with respect to the time-domain FLAFs. In particular, we have proposed a partitioned block FD-FLAF, which further reduces the latency due to the processing. We also tested several types of functional link expansions, from which we have concluded that the Chebyshev expansion is the most efficient solution when we need to work with limited resources. The proposed algorithm has been assessed in several conditions considering one of the most popular online applications that is the nonlinear acoustic echo cancellation, thus proving that even complex and powerful LIP nonlinear adaptive filters can be efficiently adopted and used for online applications. We cannot expect that FD-FLAFs perform better than time-domain FLAFs in the unlimited availability of computational resources. However, in the presence of strict computational constraints, the proposed class of filters definitely represents a suitable and reliable solution for nonlinear system modeling. 

Future works may involve advanced filtering architectures, like adaptive convex combinations of linear and nonlinear branches in the frequency domain, to control both the level of nonlinearity to model and also to optimize the tradeoff between performance and computational complexity. Implementation challenges related to functional expansions will be also investigated. Moreover, other online applications could be explored, e.g., nonlinear plant modeling, multisensor signal processing, real-time time-series analysis, on-device and edge machine learning applications, among others.
%
%
%
%
%
%
%
%
%
%
\bibliographystyle{IEEEtran}
\bibliography{FDFLAFrefs}

\end{document}